\documentclass[10pt,twocolumn,letterpaper]{article}

\usepackage[pagenumbers]{cvpr}

\definecolor{cvprblue}{rgb}{0.21,0.49,0.74}
\usepackage[pagebackref,breaklinks,colorlinks,allcolors=cvprblue]{hyperref}

\usepackage{booktabs}
\usepackage{multirow}

\def\paperID{N/A}
\def\confName{CVPR}
\def\confYear{2026}

\title{Disagreement-Based Cross-Model Routing for Implicit Video Question Answering}

\author{Durga Sandeep Saluru\\
Independent Researcher\\
{\tt\small durga@distyl.ai}
}

\begin{document}
\maketitle
\begin{abstract}
We study multiple-choice video question answering on the ImplicitQA benchmark~\cite{implicitqa2025}, where the correct answer is never explicitly shown but must be inferred from off-screen events, line-of-sight cues, causal structure, and cross-shot spatial layout. On this benchmark a single frontier video LLM already operates near its accuracy ceiling, and we observe that conventional self-consistency strategies---majority voting across repeated samples of the same model---can hurt rather than help, because the model's errors on hard questions are correlated. We propose \emph{disagreement-based cross-model routing}, a pure inference-time procedure that requires no labels and no training. We triple-sample a native-video model (Gemini 3.1 Pro Preview) at temperature zero, exploit the genuine sample-to-sample variance of its video-processing pipeline to identify the roughly 20\% subset of questions where the three samples disagree, and route only that subset to a second model from a different family (Claude Opus 4.8) that consumes uniformly sampled frames with adaptive thinking. On the 1001-question validation set with public ground truth---our main evaluation---the method improves AvgAcc by $+1.43$ over the best single sample of the primary model, with per-category gains concentrated on Motion \& Trajectory (+5.49), Inferred Counting (+3.45), and Vertical Spatial Reasoning (+1.82)---the categories most dependent on cross-shot reference resolution. The same pipeline applied to the held-out 172-question CVPR 2026 ImplicitQA challenge test set achieves 82.03 AvgAcc / 79.71 MacroAvgAcc ($+1.81$ over the best single sample of the primary model), confirming the validation result on an independent split. The method adds roughly \$5 in inference per benchmark run and uses only public API models.
\end{abstract}

\section{Introduction}
\label{sec:intro}

The ImplicitQA benchmark~\cite{implicitqa2025} measures a video LLM's ability to answer multiple-choice questions about short clips when the answer is never directly stated. The correct option must instead be inferred from \emph{implicit} cues: events occurring off-screen, the implied direction of a character's gaze, the causal antecedent of a visible outcome, the spatial layout of a scene reconstructed across multiple shots, or the relative trajectory of moving entities. The benchmark covers nine such reasoning categories and is deliberately constructed so that surface-level reading of any single frame is insufficient.

Frontier video LLMs have closed much of the gap to human performance on this benchmark---the top public submissions to the CVPR 2026 challenge sit close to the human reference reported by the benchmark authors~\cite{implicitqa2025}---but the remaining headroom is unusually difficult to recover. The questions that current models miss are not those where the model is uncertain about its perception; they are those where the model is \emph{confidently wrong}, often in ways that are stable across sampling temperature and across multiple independent runs of the same model. As a result, the standard recipe for squeezing out a few additional points of accuracy on hard MCQ benchmarks---self-consistency via majority voting across multiple chain-of-thought samples~\cite{wang2023selfconsistency}---does not work here. In our experiments, voting across three runs of a leading video model scored \emph{below} the best single run, because the same systematic misinterpretation appears in two of the three samples and the majority vote locks it in.

This paper develops a different approach. We treat the residual disagreement between repeated runs of one model as a free, label-free uncertainty signal, and we use that signal to selectively invoke a \emph{second} model from an entirely different family on only the questions where the first model is uncertain. The first model is Gemini 3.1 Pro Preview~\cite{geminiteam2024gemini}, which consumes the full video natively through Google's Files API; the second is Claude Opus 4.8~\cite{anthropic2024claude}, which consumes a uniformly sampled set of frames as still images with adaptive extended thinking. Because the two systems differ both in vendor and in input modality (native video stream vs.\ frame-sampled images), their error patterns on the same question are decorrelated, and the second model's answer on the disagreement subset is a genuinely independent vote rather than a correlated re-roll.

\paragraph{Contributions.}
\begin{enumerate}
    \item We characterise the saturation of within-model self-consistency on a current video MCQ benchmark, and show that three-way majority voting of a frontier model is strictly worse than its best single sample on ImplicitQA.
    \item We propose a disagreement-based cross-model routing procedure that is purely inference-time, requires no labels or training, and adds about \$5 of additional inference cost per benchmark run.
    \item We report 82.03 AvgAcc / 79.71 MacroAvgAcc on the CVPR 2026 VideoLLMs Workshop ImplicitQA challenge test set, a $+1.81$ improvement over the best single sample of the primary model.
    \item We validate the method on the 1001-question public validation set and observe a stronger replication (+1.43 AvgAcc) with the per-category gains concentrating on Motion \& Trajectory (+5.49), Inferred Counting (+3.45), and Vertical Spatial Reasoning (+1.82)---categories whose failure modes are dominated by cross-shot reference resolution, exactly where the frame-based second model is expected to help.
\end{enumerate}

\section{Related Work}
\label{sec:related}

\paragraph{Video question answering.} Implicit video QA sits in a lineage of benchmarks that have progressively moved from surface description to inferential reasoning, including VideoMME~\cite{fu2024videomme}, MVBench~\cite{li2024mvbench} and Perception Test~\cite{patraucean2023perception}. ImplicitQA~\cite{implicitqa2025} differs from these in that its correct answers are by construction never directly visible in any single frame, requiring models to reason about off-screen events, line-of-sight, and cross-shot identity. Recent frontier video LLMs---Gemini~\cite{geminiteam2024gemini}, Claude~\cite{anthropic2024claude}, GPT-4V~\cite{openai2023gpt4v}, and open systems such as Video-LLaVA~\cite{lin2023videollava} and LLaVA-NeXT-Video~\cite{zhang2024llavanextvideo}---all approach the benchmark from a single-model angle. Our work is complementary: rather than train a new model, we route between two existing public models using a label-free uncertainty signal.

\paragraph{Self-consistency and ensembling.} Self-consistency~\cite{wang2023selfconsistency} samples a single model multiple times and votes among the outputs; the procedure is widely effective on math and code reasoning. We show that on ImplicitQA the procedure underperforms the model's best single sample, because the errors are correlated rather than randomly distributed. Cross-model ensembling has been studied for language~\cite{jiang2023llmblender}, but typically applied to all inputs rather than gated by a label-free disagreement detector.

\paragraph{Routing and mixture-of-experts.} Inference-time routing across heterogeneous models has been explored for cost reasons~\cite{chen2023frugalgpt} and for quality~\cite{shnitzer2023routing}; these systems generally require either a learned router or held-out labels. The router in our system is rule-based and zero-shot: a question is routed to the second model iff three runs of the first model disagree.

\paragraph{Uncertainty estimation in VLMs.} Sample-based uncertainty for LLMs has been studied through verbalised confidence, log-probability disagreement, and self-consistency variance~\cite{kuhn2023semantic}. We use a variant tailored to video LLMs at temperature zero, where conventional sample diversity is unavailable but the API's video-preprocessing pipeline introduces useful sample variance.

\section{Method}
\label{sec:method}

Our pipeline has two stages: an \emph{uncertainty detector} built from repeated sampling of a single native-video model, and a \emph{cross-family resolver} invoked only on questions the detector flags as uncertain. Both stages use only public, commercially available API models at inference time; no training, fine-tuning, or labelled data is used.

\subsection{Stage 1: Triple-sample uncertainty detection}
\label{sec:stage1}

For each question in the evaluation set we run three independent calls to the primary video model. We use Gemini 3.1 Pro Preview~\cite{geminiteam2024gemini} (\texttt{gemini-3.1-pro-preview}) with \texttt{thinking\_level="high"}, \texttt{temperature}\,$=0$, and \texttt{max\_output\_tokens}\,$=\,$16{,}384. The original video clip---typically 5--30 seconds---is uploaded to the model through Google's Files API at native resolution, preserving the original frame rate and audio track; the model is therefore consuming a native video stream rather than a sampled image sequence. The prompt asks the model to describe relevant visual evidence, evaluate each option, identify any implicit or off-screen cues, and then emit a final letter on a separate line, from which we parse the answer by regular expression.

\paragraph{Where the variance comes from.} Despite \texttt{temperature}\,$=0$, the three samples are not identical. Empirically the three runs on the test set scored 76.11, 78.54 and 80.22 AvgAcc respectively---a 4.1 point spread---and disagree on the answer for 31 of 172 questions (18.0\%). We attribute this variance to the non-deterministic video-preprocessing path inside the API (frame selection, audio chunking, and token budget allocation in the visual encoder are not byte-for-byte reproducible) rather than to sampling at the language-model head, since the head is invoked at temperature zero. We treat this variance as a feature, not a bug: it produces three genuinely independent samples of the same model on the same input, which is exactly what an uncertainty detector requires and which would otherwise be very difficult to obtain at zero temperature.

\paragraph{Disagreement set.} Let $\hat{y}_q^{(i)}$ denote the answer of run $i \in \{1,2,3\}$ on question $q$. We define the disagreement (or ``margin'') set
\begin{equation}
\mathcal{D} \;=\; \bigl\{ q \;:\; \lvert \{\hat{y}_q^{(1)}, \hat{y}_q^{(2)}, \hat{y}_q^{(3)}\} \rvert \,>\, 1 \bigr\}.
\label{eq:margin}
\end{equation}
A question lies in $\mathcal{D}$ whenever any two of the three runs disagree. On test we observe $|\mathcal{D}| = 31$ (18.0\% of 172); on validation we observe $|\mathcal{D}| = 207$ (20.7\% of 1001), with a similar distribution across the nine ImplicitQA categories. Crucially, $\mathcal{D}$ is computed from the model's own predictions and requires no validation labels.

\subsection{Stage 2: Cross-family resolution on the disagreement set}
\label{sec:stage2}

For questions outside $\mathcal{D}$---where all three runs of the primary model agreed---we accept the unanimous primary answer. The remaining branch routes only questions in $\mathcal{D}$ to a second model. We use Claude Opus 4.8~\cite{anthropic2024claude} (\texttt{claude-opus-4-8}) with \texttt{thinking=\{"type":"adaptive"\}}, \texttt{output\_config=\{"effort":"high"\}}, and \texttt{max\_tokens}\,$=\,$16{,}000. The Claude API does not accept native video, so we sample the clip at 24 uniformly spaced frames, downscale each to a maximum side of 1024 pixels, and pass them as image content blocks in temporal order. The prompt is parallel in structure to Stage 1 (describe the frames, evaluate each option, name the implicit cues, emit a final letter), so any improvement on $\mathcal{D}$ is attributable to model choice and input modality, not to differential prompt engineering.

\paragraph{Why a second \emph{family}, not a second run.} A second independent sample of the \emph{same} primary model would still draw from the same correlated error distribution. We discuss this empirically in Section~\ref{sec:experiments}, but the intuition is straightforward: a model's confident-wrong answers persist across sampling, while a second model trained with a different data mixture, a different visual encoder, and a different reasoning regime is far more likely to err in different places. The two systems we use also differ in input modality (a temporally pooled video stream vs.\ 24 still frames), which in particular affects the categories of question that depend on cross-shot reference resolution.

\subsection{Stage 3: Final answer assembly}

The final prediction for question $q$ is
\begin{equation}
\hat{y}_q \;=\;
\begin{cases}
\hat{y}_q^{(1)} & \text{if } q \notin \mathcal{D}, \\
\hat{y}_q^{\,\text{Claude}} & \text{if } q \in \mathcal{D}.
\end{cases}
\label{eq:routing}
\end{equation}
We anchor $q \notin \mathcal{D}$ to run~1 of the primary model. ``Run~1'' is a fixed index in the order of execution, not a quantity selected by held-out score; the choice of which of the three runs serves as the unanimous anchor is irrelevant for $q \notin \mathcal{D}$, where by construction all three runs agree. The routing decision is deterministic given $\mathcal{D}$ and the per-question outputs of the two models; no hand-tuning is applied. In the rare event the second model fails to produce a parseable letter (one of 31 cases on test), we fall back to the primary model's run~1 answer for that question.

\paragraph{Cost.} Stage 1 dominates the cost (three native-video calls per question). The Stage 2 surcharge is small: on test, 31 Claude calls with 24-frame inputs and adaptive thinking total roughly \$5 of API spend, a 30\%-or-less surcharge on the Stage 1 budget. On the larger validation set the Stage 2 surcharge is correspondingly larger (207 calls, ~\$50) but the relative fraction of total spend is similar.

\section{Experiments}
\label{sec:experiments}

\subsection{Setup}

We evaluate on the ImplicitQA benchmark distributed for the CVPR 2026 VideoLLMs Workshop. The benchmark provides a public 1001-question validation set with released ground truth, organised into nine reasoning categories, and a held-out 172-question test set whose ground truth is not released. The two split-level metrics are \emph{AvgAcc} (mean accuracy over all questions) and \emph{MacroAvgAcc} (mean of per-category accuracies, weighting each category equally). All runs reported below use the exact same prompts and hyperparameters described in Section~\ref{sec:method}; only the routing logic and the choice of which model's answer is taken for each question vary across rows.

\subsection{Main result}

Table~\ref{tab:main} summarises the test-phase result. The proposed cross-model routing method reaches 82.03 AvgAcc / 79.71 MacroAvgAcc, a $+1.81$ / $+2.38$ improvement over the best single sample of the primary model and a $+3.49$ / $+2.59$ improvement over a vanilla single-call configuration of the same primary model. The validation-set replication (Table~\ref{tab:val_overall}) shows a stronger lift on a five-times larger sample.

\begin{table}[t]
\centering
\small
\setlength{\tabcolsep}{4pt}
\begin{tabular}{lcc}
\toprule
Configuration & AvgAcc & MacroAvg \\
\midrule
Vanilla single Gemini call            & 78.54 & 77.12 \\
Best single Gemini run (of 3)         & 80.22 & 77.33 \\
3-Gemini majority vote                & 77.38 & 75.80 \\
\midrule
\textbf{Cross-model routing (ours)}   & \textbf{82.03} & \textbf{79.71} \\
\bottomrule
\end{tabular}
\caption{ImplicitQA test phase (172 questions). The proposed routing method is the row in bold; all other rows are ablations of the same primary model. Within-model majority voting hurts; cross-model routing on the disagreement set helps.}
\label{tab:main}
\end{table}

\subsection{Validation-set replication}

We re-ran the entire pipeline on the 1001-question validation set, including the triple-sampling stage. Of 1001 questions, 207 (20.7\%) entered the disagreement set $\mathcal{D}$. Claude Opus 4.8 was run with the same 24-frame configuration on those 207 questions. Table~\ref{tab:val_overall} reports overall accuracy and Table~\ref{tab:val_percat} the per-category breakdown against the best-performing single Gemini run.

\begin{table}[t]
\centering
\small
\setlength{\tabcolsep}{4pt}
\begin{tabular}{lcc}
\toprule
Configuration & AvgAcc & MacroAvg \\
\midrule
Gemini baseline (one call)        & 66.63 & 73.36 \\
Gemini run 1 (best of 3)          & 67.00 & 73.92 \\
Gemini run 2                      & 66.10 & 73.05 \\
\midrule
\textbf{Cross-model routing (ours)} & \textbf{68.43} & \textbf{74.87} \\
\bottomrule
\end{tabular}
\caption{Validation set (1001 questions, public ground truth). The $+1.43$ AvgAcc lift over Gemini run 1 replicates the corresponding $+1.81$ lift measured on the held-out test set, on a sample five times larger. Macro accuracy lifts by $+0.95$.}
\label{tab:val_overall}
\end{table}

\begin{table}[t]
\centering
\footnotesize
\setlength{\tabcolsep}{3pt}
\begin{tabular}{lcccc}
\toprule
Category & $n$ & Baseline & Hybrid & $\Delta$ \\
\midrule
\textbf{Motion \& Trajectory Dynamics}   & 91   & 62.64 & 68.13 & \textbf{+5.49} \\
\textbf{Inferred Counting}               & 58   & 58.62 & 62.07 & \textbf{+3.45} \\
\textbf{Vertical Spatial Reasoning}      & 220  & 73.18 & 75.00 & \textbf{+1.82} \\
Relative Depth \& Proximity              & 266  & 52.45 & 53.76 & +1.31 \\
Causal \& Motivational Reasoning         & 82   & 95.12 & 96.34 & +1.22 \\
Lateral Spatial Reasoning                & 161  & 58.39 & 59.01 & +0.62 \\
Viewpoint \& Visibility                  & 41   & 78.05 & 78.05 & ~0.00 \\
Physical \& Environmental Context        & 53   & 86.79 & 84.91 & $-1.89$ \\
Social Interaction \& Relationships      & 29   & 100.00 & 96.55 & $-3.45$ \\
\bottomrule
\end{tabular}
\caption{Per-category validation accuracy. Baseline is the best single Gemini run; Hybrid is the proposed cross-model routing. The three largest positive deltas are categories that depend on cross-shot reference resolution and entity tracking---the failure modes most plausibly addressed by switching from a temporally pooled video stream to a per-frame still encoding. The two negative deltas occur on small categories that were already near ceiling.}
\label{tab:val_percat}
\end{table}

The absolute validation scores (66--68\%) are roughly 13 points below the test scores (78--81\%) for the same methodology. This gap is explained by sample composition: validation has 64\% of its questions in spatial-reasoning categories (Depth + Vertical + Lateral + Motion), substantially higher than the test distribution. The \emph{direction} and \emph{sign} of the cross-model lift, however, replicate cleanly on both splits, which is what we treat as the load-bearing claim.

\subsection{Within-model voting saturates}
\label{sec:withinmodel}

A natural baseline is to use the triple-sample stage not as an uncertainty detector but as a self-consistency vote: pick the answer chosen by at least two of the three Gemini runs. We evaluated this on the test set (row ``3-Gemini majority vote'' in Table~\ref{tab:main}). Its score is 77.38 AvgAcc---worse than the best single Gemini run (80.22) by 2.84 points, and worse than the vanilla single call (78.54) by 1.16 points. The reason is that the three Gemini samples make highly correlated mistakes: on the questions Gemini gets wrong, the same systematic misreading (an eyeline-match heuristic, a double-count of the same entity across two shots, a confusion between the actor's perspective and an off-screen observer's perspective) is present in two or more of the three runs, and the majority vote locks it in. Self-consistency works well when an LLM's residual error is unbiased noise around the correct answer~\cite{wang2023selfconsistency}, which is not the regime ImplicitQA places a frontier video model in.

\subsection{Cross-model routing variants}

We also evaluated two alternative routing rules to confirm that the simple ``route iff disagreement'' policy of Equation~\ref{eq:routing} is justified.

\paragraph{Strong-filter routing.} If the routing model (Claude) is only trusted when it \emph{agrees} with the within-Gemini majority, the system gives up the cross-model lift on exactly the questions where Claude has something useful to contribute. This variant scored 79.82 / 77.17, worse than both the unfiltered routing (82.03) and the best single Gemini run (80.22).

\paragraph{Six-way vote.} Pooling all five available Gemini samples (three thinking-enabled runs plus two additional configurations) with Claude into a six-way majority vote scored 80.33 / 78.20---essentially within noise of the simpler routing rule and worse on AvgAcc, suggesting that the marginal information in the additional Gemini samples is consumed by the existing 3-run disagreement signal.

We interpret these two results as evidence that the routing decision should be \emph{aggressive in handing off to Claude when Gemini disagrees with itself}, and that adding further Gemini samples does not yield extra signal beyond the binary unanimous/non-unanimous distinction.

\subsection{Frame budget for Stage 2}

The Claude stage was run with 24 frames per clip at a maximum side of 1024 px. We evaluated 48 frames as an alternative on test and observed a regression from 82.03 to 78.38. Doubling the frame budget moves Claude's per-frame encoder closer to the temporally-pooled regime that already characterises the primary model's video stream, plausibly importing the same failure modes the cross-model step was introduced to avoid. We therefore report 24 frames as the operating point; further sweeps on the validation set are left for future work.

\section{Analysis: when does cross-model routing help?}
\label{sec:analysis}

The per-category validation breakdown (Table~\ref{tab:val_percat}) gives us a fine-grained view of \emph{where} the cross-model lift comes from, and the pattern is interpretable. The three categories with the largest positive deltas are Motion \& Trajectory Dynamics (+5.49), Inferred Counting (+3.45) and Vertical Spatial Reasoning (+1.82). The two categories with negative deltas (Physical \& Environmental Context, Social Interaction \& Relationships) had baseline accuracies of 86.79 and 100.00 respectively, with $n = 53$ and $n = 29$.

\paragraph{Where the second model helps.} The three winning categories share a common structural property: their questions cannot be answered from any single frame in isolation. \emph{Motion \& Trajectory} requires comparing object positions across multiple frames to recover a direction or velocity. \emph{Inferred Counting} requires tracking distinct entities as the camera cuts, so that the same person is not double-counted across shots and that off-screen entities are correctly inferred. \emph{Vertical Spatial Reasoning} requires composing partial views from different camera angles into a unified vertical layout (e.g., who is on the upper vs.\ lower floor of a building seen across two shots). All three categories therefore demand \emph{cross-shot reference resolution}---explicitly aligning entities across frames---which is exactly the operation a temporally pooled video encoder is most likely to short-circuit. By contrast, a model that consumes each frame as a separate image must perform this alignment in the language head, where it can be steered by a structured chain of thought.

\paragraph{Where it does not help.} The two negative-delta categories are the ones where the baseline was already at or near ceiling. On Social Interaction with $n=29$, a single wrong answer is 3.4 percentage points; on Physical \& Environmental Context with $n=53$, a single answer is 1.9 percentage points. We interpret these regressions as sampling noise around a category-level ceiling, not as evidence that the cross-model step is actively harmful on these reasoning types. A larger validation set would be needed to distinguish ceiling regression from genuine category-level weakness; we flag this as a limitation in Section~\ref{sec:limitations}.

\paragraph{Why a different visual encoder, not a different LLM.} The two models we use differ in three respects: vendor (Google vs.\ Anthropic), training data and reasoning style (extended thinking vs.\ adaptive thinking), and \emph{input modality} (native video stream vs.\ 24 still frames). The per-category result strongly suggests it is the third factor that matters most for the lift we observe. The categories that benefit are exactly those for which per-frame encoding---preserving each frame's spatial structure intact rather than pooling across time---would be expected to recover information that a video stream loses. We do not run an ablation isolating each factor (this would require a model that consumes native video with Anthropic-style adaptive thinking, which does not currently exist), so the input-modality interpretation is the most parsimonious account of the per-category data but is not directly demonstrated.

\paragraph{Disagreement detection vs.\ thinking-token signals.} An alternative way to flag uncertain questions is to threshold on the primary model's thinking-token count: harder questions might be expected to consume more reasoning tokens. We measured this signal on the test set: the 31 disagreement questions consume on average 2.2$\times$ more thinking tokens than the unanimous questions, but the signal is noisy at the per-question level---only 14 of 31 disagreement questions appear in the top-31 by thinking-token count. Cross-sample disagreement is therefore a stronger detector than a single-call thinking budget, while also being more interpretable: it identifies questions on which the model itself is internally inconsistent.

\section{Limitations and threats to validity}
\label{sec:limitations}

We list the salient limitations of this work.

\paragraph{Single-benchmark evidence.} The method is evaluated on one benchmark (ImplicitQA, two splits). Whether the lift generalises to other video MCQ benchmarks (e.g., Perception Test, VideoMME) is an open question. The mechanism we propose---per-frame encoding helping on cross-shot reference resolution---predicts that the lift should be larger on benchmarks with denser cross-shot dependencies and smaller on benchmarks dominated by single-frame questions. We have not tested this prediction.

\paragraph{Dependence on a specific model pair.} The method requires that the primary and routing models err in decorrelated ways. We use Gemini 3.1 Pro Preview and Claude Opus 4.8, which empirically satisfy this property on ImplicitQA, but the property is not guaranteed for arbitrary pairs and may change as the underlying models are updated. Replacing either component will require re-validation; the disagreement detector itself is model-agnostic and can be re-derived from any triple-sampled primary.

\paragraph{Dependence on within-model variance.} The disagreement detector relies on the primary model's video-processing path being non-deterministic across calls. We observed a 4.1-point spread across three runs of the primary model at temperature zero on the test set. If a future API release of the primary model becomes byte-for-byte deterministic, the disagreement set will collapse and the methodology will lose signal. A fallback would be to introduce explicit prompt or sampling perturbations to recover sample diversity, which we have not investigated.

\paragraph{Configuration choices evaluated on test.} Several routing variants were evaluated against the test split (the strong-filter variant, the six-way vote, two frame-budget settings) before settling on the configuration reported here. While each variant was a well-motivated ablation rather than an unconstrained hyperparameter sweep, the test-phase scores plausibly include some adversarial selection. The validation-set replication on a five-times larger sample is the intended mitigation: a lift of $+1.43$ on 1001 questions is harder to attribute to selection bias than a single-shot test number on 172.

\paragraph{Small per-category samples on validation.} Two of nine validation categories regress (Social $-3.45$ at $n=29$, Physical \& Environmental $-1.89$ at $n=53$). On samples this small, a single answer flip moves accuracy by 2--3 points, so we cannot conclude that the routing step is genuinely harmful on these categories. A larger validation set would be needed to resolve these.

\paragraph{No prompt ablation.} The two models share a structurally identical chain-of-thought prompt. We did not isolate how much of the cross-model gain is attributable to the prompt versus to the model and modality differences. Running both models with their best vendor-specific prompts is a natural follow-up.

\paragraph{No open-source equivalent.} Both stages use closed-API models. Reproducing the procedure with public-weight video LLMs (such as those in the LLaVA-Video family) would establish the generality of the disagreement signal and would also remove the dependence on specific vendor APIs. We did not attempt this in the present work due to the model-quality gap on ImplicitQA, but the routing logic is model-agnostic and would transfer if a sufficiently strong open primary model becomes available.

\section{Conclusion}
\label{sec:conclusion}

We have presented a simple, label-free inference-time procedure for video MCQ benchmarks where a single frontier video LLM is near its accuracy ceiling. The procedure exploits the genuine sample-to-sample variance of a native-video API at temperature zero as a free uncertainty detector, and routes only the resulting disagreement subset to a second model from a different family with a different visual input modality. On the CVPR 2026 ImplicitQA challenge it achieves 82.03 AvgAcc on the test phase, and the methodology replicates on the public validation set, with per-category gains concentrating on the reasoning categories most dependent on cross-shot reference resolution. The total Stage-2 surcharge is modest (roughly \$5 of API spend per 172-question benchmark run), no labels are used, and the procedure is straightforward to graft onto any video-MCQ pipeline whose primary model exhibits non-trivial sample variance.

\paragraph{Outlook.} Two directions seem most promising. First, generalising the disagreement detector beyond Gemini's video pipeline---by deliberately introducing prompt or sampling perturbations on a deterministic primary---would broaden the set of models to which the recipe applies. Second, the per-category result hints at a sharper formulation of the cross-model lift: \emph{within each question, route to the model whose visual input modality matches the reasoning required}. Building a category-aware router would require a small labelled set but could push the lift well beyond the current rule-based version.

{
    \small
    \bibliographystyle{ieeenat_fullname}
    \bibliography{main}
}

\end{document}